\documentclass{article}
\usepackage{amsmath}

\PassOptionsToPackage{numbers, compress}{natbib}
\usepackage[preprint]{neurips_2026}

\usepackage[utf8]{inputenc} 
\usepackage[T1]{fontenc}    
\usepackage{hyperref}       
\usepackage{url}            
\usepackage{booktabs}       
\usepackage{amsfonts}       
\usepackage{nicefrac}       
\usepackage{microtype}      
\usepackage{xcolor}         
\usepackage{multirow}
\usepackage{tikz}

\usepackage{pgfplots}
\usetikzlibrary{arrows.meta,calc,positioning,fit,decorations.pathreplacing}
\pgfplotsset{compat=1.18}

\title{Crowded in B-Space: Calibrating Shared Directions for LoRA Merging}

\author{Yixuan Tang, Yi Yang \\
The Hong Kong University of Science and Technology\\
\texttt{ytangch@connect.ust.hk, imyiyang@ust.hk}
}

\begin{document}

\maketitle

\begin{abstract}
    Merging separately trained LoRA adapters is a practical alternative to joint multi-task training, but it often hurts performance. Existing methods usually treat the LoRA update $\Delta W = BA$ as a single object and do not distinguish the two LoRA matrices. We show that the main source of LoRA merge interference comes from the output-side matrix $B$. Across tasks, $B$ repeatedly uses a small set of shared directions, while $A$ remains much more task-specific. As a result, the merged adapter overemphasizes these shared directions, and task-specific information is lost. We propose \textbf{Pico} (\textbf{P}re-merge \textbf{i}nterference \textbf{c}alibration in \textbf{o}utput-space), a data-free method that calibrates $B$ before merge by downscaling over-shared directions and then rescaling the merged update. Pico plugs directly into existing merging methods such as Task Arithmetic, TIES, and TSV-M. Across eight different benchmarks from math, coding, finance, and medical domains, Pico improves average accuracy by 3.4--8.3 points over the corresponding base method and achieves the best overall average performance. Pico also enables merged adapters to outperform the LoRA trained with all task data. These results show that LoRA merging works better when the two LoRA matrices are treated separately.
\end{abstract}

\section{Introduction}
Fine-tuning pretrained models for different tasks, domains, or data distributions has become standard practice, producing many specialized checkpoints from the same base model. However, leveraging these diverse checkpoints effectively remains challenging.  Model merging offers a promising solution by combining them into a single model without joint retraining~\citep{ilharco2023ta}.  It is now widely used, from composing skills across tasks~\citep{ilharco2023ta,stoica2025knots} to averaging training checkpoints for improved robustness~\citep{modelsoup,qwen3embedding,geminiembedding}.

Low-Rank Adaptation (LoRA)~\citep{hu2022lora} parametrizes each task-specific update as a small low-rank adapter. Because adapters are compact and easy to distribute, thousands of them are now publicly available on platforms such as Hugging Face~\citep{prabhakar2025lorasoup,zhao2025loralego}. Merging these adapters is a natural way to combine their capabilities. Yet merging independently trained adapters often degrades their individual performance. To mitigate this issue, existing methods attempt to reduce  merge interference by resolving sign conflicts~\citep{yadav2023tiesmerging}, pruning small weights~\citep{dare,deep2024della}, or aligning updates in a shared basis~\citep{stoica2025knots}. However, these methods all treat the LoRA update $\Delta W = BA$ as one object. They do not distinguish the two LoRA matrices that produce it, i.e., $A$ and $B$.


This matters because the two LoRA matrices play different roles. In $\Delta W = BA$, $A$ maps input features into a low-rank space, while $B$ maps that representation back to the output space. Prior work already suggests that finetuning is asymmetric across the two matrices. \citet{pmlr-v235-zhu24c} show that, in single-task finetuning, a larger share of the update change is carried by $B$. Practical methods exploit the same asymmetry. LoRA-FA~\citep{zhang2023lorafa} freezes $A$ after initialization to reduce memory, and HydraLoRA~\citep{tian2024hydralora} shares a single $A$ across task-specific $B$ heads. These designs treat $A$ as the easier factor to freeze or share, while leaving more task-specific change in $B$.


\begin{figure*}[tbp]
    \centering
    \includegraphics[width=\linewidth]{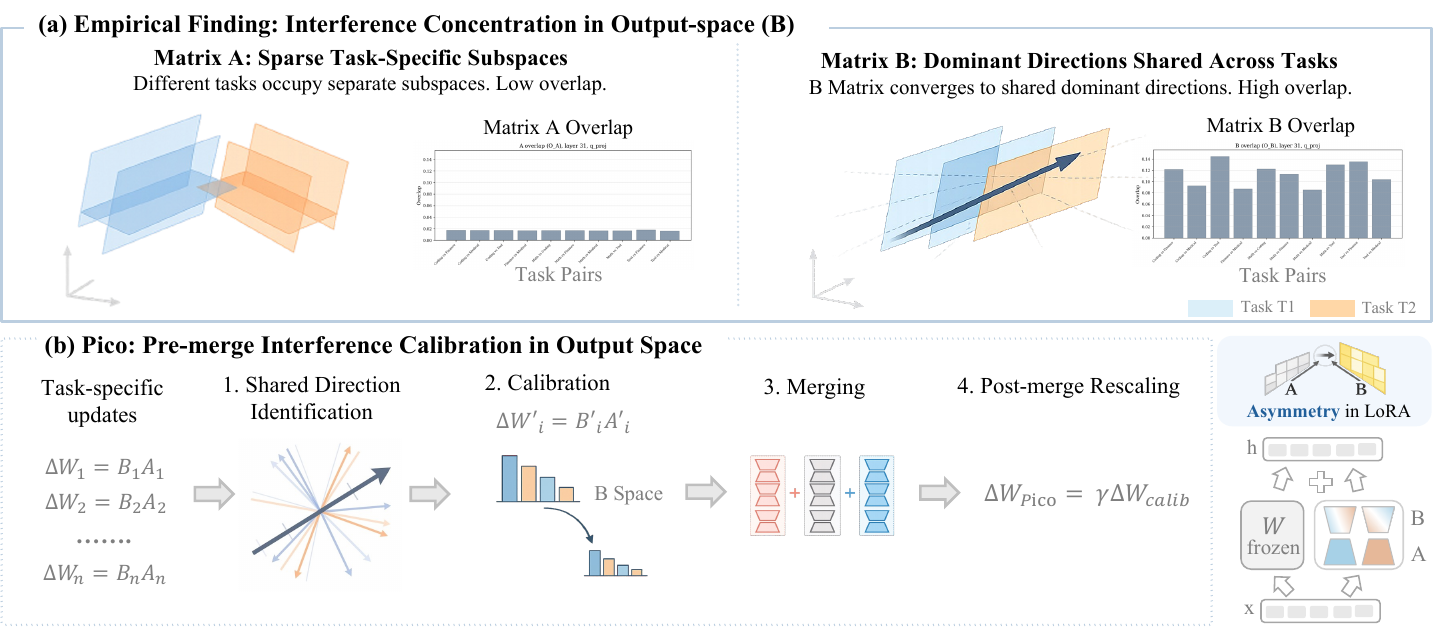}
    \caption{\textbf{Overview of Pico.} (a) Merging interference in LoRA is asymmetric: task-specific $A$ matrices remain relatively separated, while $B$ matrices align along shared dominant directions across tasks. (b) Pico identifies and downscales these shared directions in $B$ before merging, then rescales the merged update to preserve its magnitude.}
    \label{fig:main}
\end{figure*}

This creates a natural question for merging. Existing merge methods still operate on the full update $\Delta W = BA$, even though the two factors do not contribute equally during finetuning. If $B$ carries more of the adapted change, it may also carry more of the merge interference when independently trained adapters are combined. We therefore ask: \textit{does merge interference come from the full update, or mainly from the output-side matrix $B$?}

Our answer is the output-side matrix $B$. Across independently trained LoRAs for math, coding, finance, and medical reasoning, cross-task overlap is consistently much larger in $B$ than in $A$, and this gap grows with rank. At the same time, the effective rank of $B$ stays low even at higher ranks, so different tasks keep reusing a small set of output-space directions. Together, these two patterns point to a simple failure mode: the main problem is not uniform conflict across the full update, but repeated accumulation of a few shared directions in $B$.

This suggests that the fix should also focus on $B$. We propose \textbf{Pico} (\textbf{P}re-merge \textbf{i}nterference \textbf{c}alibration in \textbf{o}utput-space), a data-free calibration step applied to task-specific $B$ matrices before merging. Pico identifies shared directions in $B$, downscales them according to how strongly they align across tasks, and then rescales the merged update to preserve its overall magnitude. Pico is not a replacement merger. Instead, it plugs directly into methods such as Task Arithmetic~\citep{ilharco2023ta}, TIES~\citep{yadav2023tiesmerging}, and TSV-M~\citep{tsvm}. Figure~\ref{fig:main} gives an overview of the problem and the method.

Experiments across eight datasets from four domains show that this simple calibration consistently improves downstream merging. Pico improves over the corresponding uncalibrated baselines by 3.4 points for Task Arithmetic, 4.7 points for TIES, and 8.3 points for TSV-M. It also outperforms DARE~\citep{dare}, DELLA~\citep{deep2024della}, KnOTS~\citep{stoica2025knots}, and Core Space~\citep{panariello2025core} in overall average performance, and enables merged adapters to outperform a LoRA adapter trained jointly on all task data. Overall, this paper makes three contributions. First, it reveals that LoRA merge interference is structurally concentrated in the output-side matrix $B$ rather than uniformly distributed across the full update $\Delta W$. Second, it traces this failure mode to repeated accumulation of a few shared directions in $B$ across tasks. Third, it introduces a simple plug-in calibration step that corrects this over-accumulation and improves a range of existing LoRA merging methods.

\section{Background and Problem Setup}
\label{sec:background}

\paragraph{LoRA adapter merging.}
We use $t$ to index task-specific adapters. For adapter $t$, the update at a given layer is $\Delta W_t = B_t A_t$, where $A_t \in \mathbb{R}^{r \times d_{\mathrm{in}}}$, $B_t \in \mathbb{R}^{d_{\mathrm{out}} \times r}$, and $r$ is the LoRA rank. For simplicity, we absorb the usual LoRA scaling factor into $B_t$ throughout the paper. Given $T$ independently trained adapters with the same dimension, LoRA merging aims to construct a single merged update
\[
\Delta W_{\mathrm{merge}} = \mathcal{M}(\Delta W_1,\ldots,\Delta W_T)
\]
that combines their capabilities without retraining on the union of task data.

\paragraph{Merge interference.}
Under a linear merge, components shared across adapters retain their full magnitude, while task-specific components are diluted. If several adapters contain large aligned directions, those directions can dominate the merged update and make task-specific directions relatively weaker. We call this effect merge interference. Appendix~\ref{app:motivation-analysis} gives a toy derivation showing that, under simple averaging, the shared-to-specific ratio grows by a factor of \(T\), so task-specific capabilities can be lost even when each source adapter learns them well.

\paragraph{Overlap and concentration.}
We use two measurements to quantify merge interference. Overlap across a pair of adapters is measured by a normalized subspace overlap score $O_B(i,j), O_A(i,j)\in[0,1]$, where larger values mean that the two adapters share more directions in the output- or input-side matrix, respectively. Let \(Q_i^B, Q_j^B\) be orthonormal bases of the column spaces of \(B_i, B_j\) and let \(Q_i^A, Q_j^A\) be orthonormal bases of the row spaces of \(A_i, A_j\); then
\[
O_B(i,j) = \tfrac{1}{r}\bigl\|(Q_i^B)^\top Q_j^B\bigr\|_F^2,
\qquad
O_A(i,j) = \tfrac{1}{r}\bigl\|(Q_i^A)^\top Q_j^A\bigr\|_F^2.
\]
Spectral concentration is measured by effective rank, where lower values mean that fewer directions contribute substantially. 
Full definitions (including effective rank and component energy) are given in Appendix~\ref{app:motivation-analysis}. These measurements do not assume that every shared direction is harmful. Instead, they test whether shared directions accumulate unevenly across \(A\) and \(B\).


\section{Why Does LoRA Merging Fail?}

\label{sec:motivation}

\begin{figure*}[t]
    \centering
    \includegraphics[width=.82\textwidth]{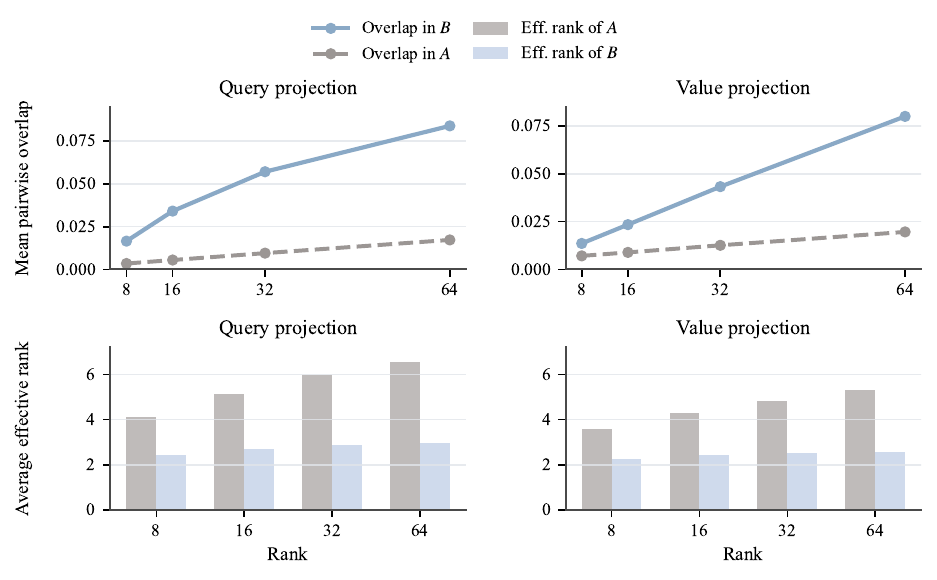}
     \caption{\textbf{Quantitative motivation for Pico.} Top row: mean pairwise overlap in $A$ and $B$ across four domain-specific LoRA adapters, shown for the query and value projection matrices. Bottom row: average effective rank of $A$ and $B$ across LoRA ranks. The exact values are reported in Appendix~A.}
    \label{fig:motivation}
\end{figure*}

\paragraph{Setup.}
We analyze four domain-specific LoRA adapters trained on Llama-3.1-8B~\citep{llama3} for math, coding, finance, and medical reasoning. To keep the comparison controlled, we sample 50k training examples per domain and fix LoRA alpha at 16. In our setup, LoRA is applied to the query and value projection matrices in attention, so we focus this analysis on those two target modules. We study LoRA ranks 8, 16, 32, and 64. Appendix~\ref{app:motivation-analysis} reports the exact datasets and full numeric results.

\paragraph{Overlap is much larger in \texorpdfstring{$B$}{B}.}
The first observation is that overlap across tasks is consistently much larger in $B$ than in $A$. At rank 8, $O_B > O_A$ already holds for 99.5\% of query-projection layer pairs and 95.3\% of value-projection layer pairs. At ranks 16, 32, and 64, it holds for every measured layer pair in both modules. The gap also grows with LoRA rank. For example, in the query projection, the mean overlap in $B$ rises from 0.0166 to 0.0839, while the overlap in $A$ grows only from 0.0035 to 0.0172. The same pattern appears in the value projection, where $B$ rises from 0.0136 to 0.0801 but $A$ increases only from 0.0071 to 0.0196. Although these scores are normalized and averaged, the asymmetry is highly consistent. Appendix~\ref{app:motivation-analysis} reports the full curves and exact values. Together, these results show that merge interference is concentrated much more strongly in $B$.

\paragraph{\texorpdfstring{$B$}{B} stays concentrated as rank grows.}
The second observation is that increasing the LoRA rank does not make $B$ much more diverse. Instead, $B$ keeps using only a small number of directions, while $A$ keeps expanding. At rank 64, the average effective rank in the query projection is 2.94 for $B$ versus 6.55 for $A$; in the value projection, it is 2.59 for $B$ versus 5.30 for $A$. Together with the overlap results, this means that independently trained adapters keep writing into the same small set of directions in $B$, while the corresponding $A$ matrices remain much more spread out. This gives a concrete picture of why naive merge degrades performance. The main problem is not uniform conflict across the full update $\Delta W = BA$. Instead, a few shared directions in $B$ are counted repeatedly during merge and end up dominating the merged adapter. This is exactly the effect that Pico targets in Figure~\ref{fig:main}(b).

\begin{figure*}[t]
\centering
\includegraphics[width=.85\textwidth]{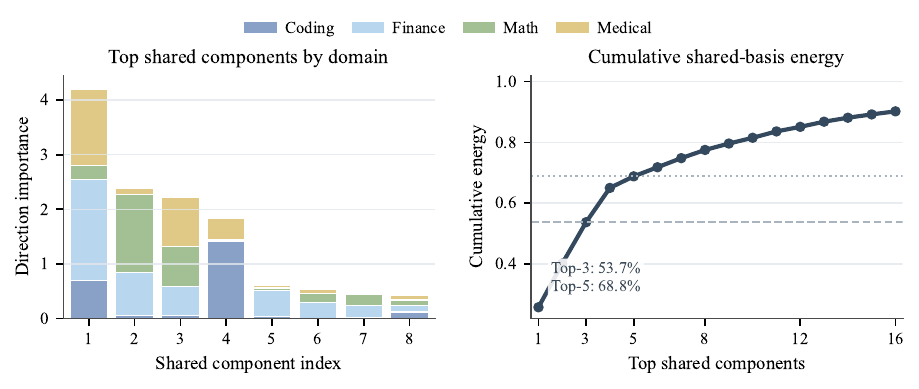}
\caption{\textbf{Representative dominance pattern in the shared $B$ spectrum} (query projection, layer 16, rank 16). Left: contribution of each task to the leading shared components. Right: cumulative energy in the same shared basis. The first five components account for 68.8\% of the total energy.}
\label{fig:coding-case-study}
\end{figure*}

\paragraph{A few shared \texorpdfstring{$B$}{B} components can dominate before merge.}
Figure~\ref{fig:coding-case-study} adds one more detail to the same diagnosis. For the query projection at layer 16, we stack the four task-specific $B$ matrices, compute a joint SVD, and inspect the leading shared components. In this example, the first three components already account for 53.7\% of the shared-basis energy, and the first five account for 68.8\%. Finance, math, and medical contribute more strongly to these dominant shared components than coding. This shows that the dominant shared $B$ components are not only concentrated, but do not represent all tasks equally before merging. When merge is dominated by a few shared $B$ components, tasks that contribute less to those components can be harder to preserve.

\section{Pico: Pre-merge interference calibration in output-space}
Pico addresses the failure mode identified in Section~\ref{sec:motivation}. For each target module in each layer, it adds one calibration step before a standard LoRA merge rule: it reduces output-side directions in \(B\) that are reused too heavily across tasks, then merges the calibrated updates and rescales the overall update magnitude. The whole procedure is data-free and plug-in: it uses only the source adapters and can be paired with different downstream merge rules.

\subsection{Shared-Direction Calibration}
\label{sec:pico-calibration}
Shared-direction calibration is the key step in Pico. As merge interference is much stronger in the output-side LoRA matrices than in the input-side matrices, Pico calibrates \(B\) before applying any downstream merge rule. For one layer, we start from task-specific updates \(\{\Delta W_t = B_t A_t\}_{t=1}^T\) and produce calibrated updates \(\{\widetilde{\Delta W}_t\}_{t=1}^T\) that are used in the next merge stage.

\paragraph{Shared Basis.}
Pico first builds a shared basis for the output space. It stacks the output-side matrices from all tasks and computes a joint singular value decomposition,
\[
B_{\mathrm{all}} = [B_1, B_2, \ldots, B_T] = U \Sigma V^\top .
\]
The columns of \(U\) define shared output-space basis vectors for this layer, and \(\sigma_j\) denotes the \(j\)-th singular value in \(\Sigma\). This gives Pico a common basis for checking how strongly different tasks use each basis component.

\paragraph{Sharing Score.}
Pico then assigns each basis component a score that measures how strongly it is shared. For the \(j\)-th basis component, we define a sharing score and a corresponding scaling coefficient,
\[
s_j = \frac{\sigma_j^2}{\sum_k \sigma_k^2},
\qquad
\alpha_j = \frac{1}{1 + (T-1)s_j}.
\]
Here \(s_j\) measures how much of the joint \(B\)-space energy lies in component \(j\), so a larger value means it is shared more heavily across tasks. The coefficient \(\alpha_j\) controls how much of that component is kept during calibration. Lightly shared components should stay near \(1\), while a component that dominates the joint spectrum should be reduced toward \(1/T\).

\paragraph{Calibration Operator.}
Pico uses these coefficients to build a layerwise calibration operator. Let \(m=\min(d_{\mathrm{out}},Tr)\), let \(\alpha = [\alpha_1,\ldots,\alpha_m]\) collect the scaling coefficients, and let \(I\) denote the identity matrix on the output space:
\[
S = I + U \,\mathrm{diag}(\alpha - 1)\, U^\top.
\]
Pico then applies this operator directly to the output-side LoRA matrix:
\[
\widetilde{B}_t = S B_t.
\]
The calibrated update is then \(\widetilde{\Delta W}_t = \widetilde{B}_t A_t\). In this way, Pico changes only the output-side matrix \(B_t\), where merge interference is strongest, while leaving the more task-specific input-side matrix \(A_t\) untouched. Computationally, each layer requires one SVD of the stacked \(B\) matrices plus one left multiplication for each task-specific \(B_t\).

\subsection{Merge and Rescaling}
\label{sec:pico-merge}
After calibration, the calibrated updates are merged with a standard LoRA merge rule:
\[
\Delta W_{\mathrm{calib}} = \mathcal{M}(\widetilde{\Delta W}_1,\ldots,\widetilde{\Delta W}_T),
\]
where \(\mathcal{M}\) can be Task Arithmetic, TIES, TSV-M, or another downstream merger. 

\paragraph{Magnitude Rescaling.}
Calibration improves the merged directions but can reduce the Frobenius norm of the merged update, leaving the adapter too weak. Pico therefore rescales the merged update to match the average magnitude of the source adapters:
\[
\Delta W_{\mathrm{Pico}} = \gamma \Delta W_{\mathrm{calib}}
\quad \mathrm{and} \quad
\gamma = \frac{1}{\|\Delta W_{\mathrm{calib}}\|_F}
\cdot
\frac{1}{T}\sum_{t=1}^T \|\Delta W_t\|_F.
\]
This scalar rescaling preserves the calibrated direction of the merged update while restoring its overall magnitude.

\section{Experiment}

We evaluate Pico in a controlled four-domain setting. We focus on two outcomes: whether it consistently improves strong LoRA merging baselines, and whether the best merged adapter remains competitive with a jointly trained multi-task LoRA.

\subsection{Experimental Setup}
\label{sec:exp-setup}

\paragraph{Training Data.}
We train four domain-specific LoRA adapters on Llama-3.1-8B~\citep{llama3}, one for each target domain. The training sets are MetaMathQA~\citep{metamathqa} for math, Magicoder-Evol-Instruct-110K~\citep{magicoder} for coding, ODA-Fin-SFT-318k~\citep{cao2026finoda} for finance, and Medical-Reasoning-SFT-Trinity-Mini~\citep{medical_reasoning_sft_trinity_mini} for medical reasoning. To keep the comparison controlled, we sample 50k training examples per domain and keep the LoRA setting fixed across domains. For the main experiments in this section, we use LoRA rank 16. In all experiments, LoRA is applied to the query and value projection matrices in attention.

\paragraph{Benchmarks.}
We evaluate merged adapters on eight benchmarks. The math domain uses GSM8K~\citep{gsmk} and MATH~\citep{math}, coding uses HumanEval~\citep{humaneval} and MBPP~\citep{mbpp}, finance uses FinanceBench~\citep{financebench} and ConvFinQA~\citep{convfinqa}, and medical uses PubMedQA~\citep{pubmedqa} and MedQA-USMLE.~\citep{medqa} We use the average score across all eight benchmarks as the main summary metric.

\paragraph{Baselines.}
Our main comparisons use three downstream merge rules: Task Arithmetic~\citep{ilharco2023ta}, TIES~\citep{yadav2023tiesmerging}, and TSV-M~\citep{tsvm}. We compare Pico against four strong baselines that reduce merge interference in different ways. DARE~\citep{dare} is a sparsification-based plug-in that drops and rescales delta parameters before parameter fusion. KnOTS~\citep{stoica2025knots} is an alignment-based method that maps LoRA updates into a shared basis before merging. DELLA~\citep{deep2024della} is a magnitude-based merging technique, and Core Space~\citep{panariello2025core} is a low-rank merging framework that also operates in a shared basis. All methods start from the same four source adapters; only the merge strategy changes. The main experiment reports domain averages. This keeps each table focused on the comparison we care about. Appendix~\ref{app:benchmark-level-results} gives the full benchmark-level results and Appendix~\ref{app:training-details} lists the training hyperparameters.


\begin{table*}[t]
\centering
\small
\setlength{\tabcolsep}{3.2pt}
\caption{\textbf{Main results across four domains.} Each domain column reports the average over its two benchmarks, and the last column reports the average over all eight benchmarks. Bold marks the best overall average in each merger block, and underlining marks the second best. Full benchmark-level results are given in Appendix~\ref{app:benchmark-level-results}.}
\label{tab:main-results}
\resizebox{\textwidth}{!}{%
\begin{tabular}{llccccc}
\toprule
Merger & Method & Math Avg. $\uparrow$ & Coding Avg. $\uparrow$ & Finance Avg. $\uparrow$ & Medical Avg. $\uparrow$ & Overall Avg. $\uparrow$ \\
\midrule
Task Arithmetic & No Calibration & 0.2666 & 0.2823 & 0.5044 & 0.5839 & 0.4093 \\
                & DARE           & 0.2757 & 0.2931 & 0.5206 & 0.5851 & \underline{0.4186} \\
                & DELLA          & 0.2651 & 0.2662 & 0.5187 & 0.5857 & 0.4089 \\
                & KnOTS          & 0.2727 & 0.2714 & 0.5127 & 0.5784 & 0.4088 \\
                & Core Space     & 0.2620 & 0.2773 & 0.4987 & 0.5880 & 0.4065 \\
                & \textbf{Pico}  & 0.3065 & 0.3372 & 0.5431 & 0.5854 & \textbf{0.4430} \\
\midrule
TIES            & No Calibration & 0.2808 & 0.2117 & 0.5271 & 0.5241 & 0.3859 \\
                & DARE           & 0.2601 & 0.1936 & 0.5130 & 0.4974 & 0.3660 \\
                & DELLA          & 0.1928 & 0.1709 & 0.4750 & 0.4025 & 0.3103 \\
                & KnOTS          & 0.2876 & 0.3012 & 0.5277 & 0.5661 & \underline{0.4207} \\
                & Core Space     & 0.2413 & 0.2490 & 0.4230 & 0.4936 & 0.3517 \\
                & \textbf{Pico}  & 0.3061 & 0.3036 & 0.5096 & 0.6116 & \textbf{0.4328} \\
\midrule
TSV-M           & No Calibration & 0.2636 & 0.1570 & 0.4487 & 0.5197 & 0.3473 \\
                & DARE           & 0.2926 & 0.1620 & 0.4481 & 0.5859 & \underline{0.3722} \\
                & DELLA          & 0.2778 & 0.1598 & 0.4662 & 0.5268 & 0.3577 \\
                & KnOTS          & 0.2666 & 0.1582 & 0.4625 & 0.5218 & 0.3523 \\
                & Core Space     & 0.2385 & 0.1415 & 0.3568 & 0.4413 & 0.2945 \\
                & \textbf{Pico}  & 0.2883 & 0.2942 & 0.5257 & 0.6139 & \textbf{0.4305} \\
\bottomrule
\end{tabular}%
}
\end{table*}

\subsection{Main Results}
\label{sec:exp-main}

\paragraph{Main finding.}
Table~\ref{tab:main-results} gives the main result. Pico improves every downstream merger we test. Relative to the corresponding no-calibration baseline, it raises the average score from 0.4093 to 0.4430 for Task Arithmetic, from 0.3859 to 0.4328 for TIES, and from 0.3473 to 0.4305 for TSV-M. These are absolute gains of 3.4, 4.7, and 8.3 points, respectively.

\paragraph{Consistency Across Mergers.}
The gains are not tied to one downstream rule. This matters because Pico is meant to correct an upstream problem in the source adapters rather than replace the merger itself. Task Arithmetic, TIES, and TSV-M make different downstream decisions, but they all start from the same source adapters. Once the over-shared directions in $B$ are calibrated before merge, all three improve.

\paragraph{Where the gains appear.}
The largest gains appear on top of TIES and TSV-M, where shared components in $B$ interact most directly with the downstream merge logic. In TIES, over-shared components survive magnitude trimming more easily and dominate the elected sign vector; in TSV-M, they concentrate energy in the low-rank singular-vector space that the merger explicitly operates on. Pico calibrates these components before either merger acts, which is why the two are complementary. Task Arithmetic also benefits, but its coordinate-level averaging is less sensitive to spectral concentration, so the gain is smaller. Overall, the same pattern repeats across all three mergers: calibrating $B$ before merge produces a more balanced adapter across domains.


\subsection{Comparison with Single-Task and Jointly Trained LoRA}
\label{sec:exp-single-joint}

\begin{table*}[t]
\centering
\small
\setlength{\tabcolsep}{3.4pt}
\caption{\textbf{Comparison with single-domain and jointly trained LoRA adapters.} Each domain column reports the average over its two benchmarks. The final row reports the best merged adapter from Table~\ref{tab:main-results}. Full benchmark-level values are given in Appendix~\ref{app:benchmark-level-results}.}
\label{tab:single-vs-joint}
\resizebox{\textwidth}{!}{%
\begin{tabular}{lccccc}
\toprule
Adapter & Math Avg. $\uparrow$ & Coding Avg. $\uparrow$ & Finance Avg. $\uparrow$ & Medical Avg. $\uparrow$ & Overall Avg. $\uparrow$ \\
\midrule
Math LoRA             & \textbf{0.2830} & 0.1090 & 0.4403 & 0.3665 & 0.2997 \\
Coding LoRA           & 0.2505 & \textbf{0.3598} & 0.4812 & 0.5516 & 0.4108 \\
Finance LoRA          & 0.2477 & 0.1440 & \textbf{0.5484} & 0.4976 & 0.3594 \\
Medical LoRA          & 0.2438 & 0.3190 & 0.4848 & \textbf{0.5950} & 0.4106 \\
\midrule
Joint Multi-Task LoRA & 0.2544 & 0.3151 & 0.4655 & 0.4405 & 0.3688 \\
\textbf{Pico w/ Task Arithmetic}
                      & 0.3065 & 0.3372 & 0.5431 & 0.5854 & \textbf{0.4430} \\
\bottomrule
\end{tabular}%
}
\end{table*}

\paragraph{Single-Domain Specialists.}
Table~\ref{tab:single-vs-joint} compares the best merged adapter from Table~\ref{tab:main-results} with four single-domain LoRAs and with a jointly trained multi-task LoRA. The single-domain adapters are highly specialized: each one performs best or near-best on its own source domain, but its average score drops on the other domains. This is the trade-off that makes adapter merging attractive in the first place.

\paragraph{Joint Training vs. Merging.}
The jointly trained multi-task LoRA is more balanced than the single-domain specialists, but its overall average is still only 0.3688. The best merged adapter, Pico with Task Arithmetic, reaches 0.4430 in Table~\ref{tab:single-vs-joint}, exceeding joint training by 7.4 points. In this setup, Pico-based merging is therefore not just competitive with joint training; it is stronger on the overall average. In this setup, calibrating the over-shared components in $B$ before merge produces a stronger multi-domain adapter than joint training on all task data.

\subsection{Ablations and Analysis}
\label{sec:exp-ablation}
We use ablations and robustness checks to answer two questions. First, is Pico's gain really tied to calibrating the output-side matrix $B$? Second, does the same advantage remain under other settings beyond the main experiment?

\begin{table*}[t]
\centering
\small
\setlength{\tabcolsep}{4.0pt}
\caption{\textbf{Calibration-space and restoration ablation on top of Task Arithmetic at rank 16.} Each domain column reports the average over its two benchmarks. Calibrating only $B$ gives the best overall result, while calibrating $A$, calibrating the full update $\Delta W$, or removing restoration lowers the overall average.}
\label{tab:calibration-space}
\resizebox{\textwidth}{!}{%
\begin{tabular}{lccccc}
\toprule
Calibration Space & Math Avg. $\uparrow$ & Coding Avg. $\uparrow$ & Finance Avg. $\uparrow$ & Medical Avg. $\uparrow$ & Overall Avg. $\uparrow$ \\
\midrule
Task Arithmetic only & 0.2666 & 0.2823 & 0.5044 & 0.5839 & 0.4093 \\
$+$ $A$-space        & 0.2485 & 0.2656 & 0.5139 & 0.5384 & 0.3916 \\
$+$ $BA$-space       & 0.2281 & 0.2378 & 0.5505 & 0.4810 & 0.3743 \\
$+$ $B$-space w/o restoration & 0.3049 & 0.2146 & 0.5646 & 0.4790 & 0.3908 \\
\textbf{$+$ $B$-space (Pico)} & \textbf{0.3065} & \textbf{0.3372} & \textbf{0.5431} & \textbf{0.5854} & \textbf{0.4430} \\
\bottomrule
\end{tabular}%
}
\end{table*}

\paragraph{Calibration space matters.}
Table~\ref{tab:calibration-space} answers two design questions. First, where should calibration be applied? Calibrating only the output-side matrix $B$ gives the best result, raising the overall average from 0.4093 to 0.4430 on top of Task Arithmetic. In contrast, calibrating $A$ or the full update $\Delta W = BA$ lowers performance. This matches the motivation analysis: the main merge interference is concentrated in $B$, while $A$ remains more task-specific. Variants that modify $A$ or the full update act on components that are not the main source of interference and can weaken useful task-specific structure.

The same table also shows that the final magnitude-restoration step is necessary after $B$-space calibration. Removing restoration drops the overall average to 0.3908 even though the calibrated directions are unchanged. The domain-level effect is not uniform: finance increases from 0.5431 to 0.5646 without restoration, but coding and medical drop sharply, and the overall result is clearly worse. In other words, direction calibration alone is not enough; the merged update also needs its magnitude to be restored.

\begin{figure*}[t]
\centering
\includegraphics[width=.95\textwidth]{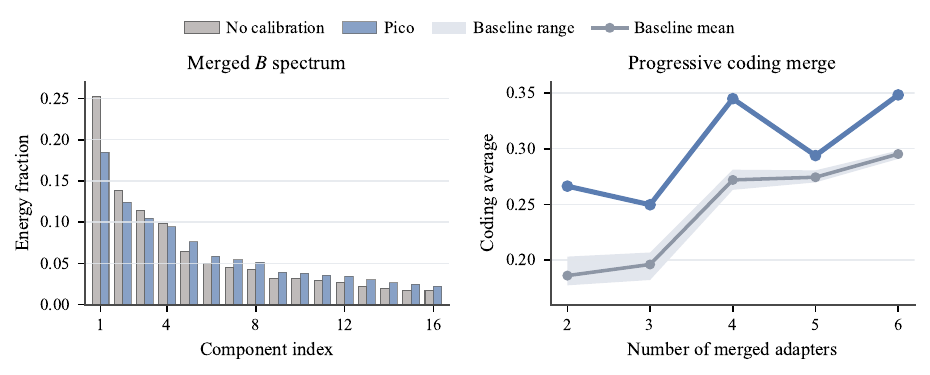}
\caption{\textbf{Spectrum calibration and progressive merge robustness.} Left: representative merged $B$ spectrum for the query projection at layer 16 with Task Arithmetic at rank 16. Pico reduces the dominance of the leading shared components. Right: coding average as the merge pool grows from two to six adapters. The band summarizes the range of non-Pico baselines; Pico remains stronger throughout the progressive merge.Appendix~\ref{app:bspace-spectrum} gives the full spectral metric definitions and exact statistics; Appendix~\ref{app:progressive-coding} gives the full progressive-merge comparison.}
\label{fig:bspace-stats}
\end{figure*}

\paragraph{The merged \texorpdfstring{$B$}{B} space becomes less skewed.}
Figure~\ref{fig:bspace-stats} shows two complementary views. The left panel is a representative layer-level spectrum of the merged output-side matrix $B$ for Task Arithmetic at rank 16. After Pico, the leading shared-basis components carry less energy, which means that the merged $B$ space is less concentrated in a few over-shared components. This is the same qualitative pattern captured by the spectral diagnostics used in Appendix~\ref{app:bspace-spectrum}: lower $o_{\max}$ means less dominance by a single component, while higher effective rank and stable rank mean that energy is spread more evenly across the shared basis. The right panel shows a harder progressive merge setting on coding benchmarks: as more adapters are added to the merge pool, Pico stays above the baseline range and keeps the coding adapter more stable.

\paragraph{Robustness beyond the main setting.}
Our main experiments focus on rank 16 on Llama-3.1-8B, but Pico’s advantage remains consistent across a broader range of settings. Appendix~\ref{app:rank-robustness} reports full results for ranks 8, 16, 32, and 64 across Task Arithmetic, TIES, and TSV-M, where Pico remains the strongest method in every setting. This again matches the motivation analysis: as the LoRA rank grows, overlap in $B$ also grows, so the merged adapter continues to face the same type of over-shared structure before merging. The effect is especially visible for TSV-M, whose no-calibration baseline drops from 0.3473 at rank 16 to 0.3206 at rank 32 and 0.2926 at rank 64, while Pico stays clearly stronger at all three ranks. Appendix~\ref{app:qwen-base-transfer} further repeats the Task Arithmetic comparison on Qwen3-4B-Base, where Pico again gives the best overall average, improving the no-calibration baseline from 0.4881 to 0.5557. 


\section{Related Work}
We position Pico against three closely related lines of work: LoRA adaptation, model and LoRA merging, and merge methods that reduce interference in shared subspaces.

\paragraph{LoRA adaptation and asymmetry.}
LoRA~\citep{hu2022lora} makes parameter-efficient finetuning practical by writing each task update as a low-rank product \(\Delta W = BA\). A growing line of work shows that these two matrices do not play symmetric roles. \citet{pmlr-v235-zhu24c} analyzes this asymmetry in single-task finetuning, and HydraLoRA~\citep{tian2024hydralora} turns the same idea into an architectural design by sharing one \(A\) across multiple task-specific \(B\) heads. Other methods also treat the two factors differently in more specific ways: LoRA-FA~\citep{zhang2023lorafa} and LoRA+~\citep{hayou2024loraplus} assign them different optimization roles, DoRA~\citep{liu2024dora} separates magnitude from direction in the adapted weights, and FedSA-LoRA~\citep{guo2025fedsalora} splits shared and client-specific updates in federated learning. Our work builds on this line of evidence, but studies a different problem: not how asymmetry affects single-task finetuning, but how it affects interference when independently trained LoRA adapters are merged.

\paragraph{Model and LoRA merging.}
Task Arithmetic~\citep{ilharco2023ta} shows that finetuned models can often be composed directly in parameter space. More recent work focuses on reducing interference during merge. TIES~\citep{yadav2023tiesmerging} trims small updates and resolves sign conflicts, DARE~\citep{dare} sparsifies and rescales delta parameters, and DELLA~\citep{deep2024della} uses magnitude-based sampling. LoRA Soups~\citep{prabhakar2025lorasoup} further frames LoRA merging as a practical tool for skill composition, while LoRA-LEGO~\citep{zhao2025loralego} pushes the granularity further by decomposing each adapter into rank-wise semantic units that can be recombined across tasks. These methods reduce merge interference in different ways, but they still operate on task updates or their coordinates rather than asking where the interference comes from inside \(\Delta W = BA\).

\paragraph{Shared-basis and low-rank merge methods.}
Pico is closest to methods that explicitly work in a shared basis or low-rank subspace. KnOTS~\citep{stoica2025knots} aligns updates in a shared singular-vector basis, TSV-M~\citep{tsvm} merges updates in a low-rank singular-vector space, and Core Space~\citep{panariello2025core} performs low-rank merging in a shared subspace. \citet{li2026sharedknowledgehurts} studies a related failure mode: aligned singular directions can be over-accumulated during linear merge, which inflates the merged singular values. Their SVC method corrects the merged spectrum afterward by rescaling those singular values. Pico differs in two ways. It traces the interference back to the LoRA decomposition itself, and it calibrates the output-side matrix \(B\) before any downstream merge rule is applied.

\section{Conclusion}
LoRA merging is a practical alternative to joint multi-task training, but existing methods usually treat the low-rank update \(\Delta W = BA\) as a single object. In this paper, we show that the main source of merge interference comes from the output-side matrix \(B\): across tasks, a few shared \(B\) components are repeatedly accumulated during merge, while \(A\) remains more task-specific. This leads the merged adapter to overemphasize those shared components and makes some task capabilities harder to preserve. Pico follows directly from this diagnosis. It calibrates only \(B\) before merge, plugs into existing merge rules, and consistently improves strong LoRA merging baselines across four domains and three downstream mergers. It also yields a stronger multi-domain adapter than a jointly trained multi-task LoRA in our setting. More broadly, these results suggest that adapter merging works better when interference is diagnosed inside the factorization rather than only at the level of the full update.

\bibliographystyle{apalike}
\bibliography{reference}

\appendix

\section{Additional Motivation Analysis}
\label{app:motivation-analysis}
This appendix supports the motivation analysis in Section~\ref{sec:motivation}. We first give a toy derivation showing how shared directions can dominate a linear merge, then define the diagnostics used in the main text, and finally report the exact statistics behind Figure~\ref{fig:motivation}. Unless otherwise noted, all measurements are computed from the four domain-specific LoRAs used in the main text: Math, Coding, Finance, and Medical. The corresponding training datasets are MetaMathQA~\citep{metamathqa}, Magicoder~\citep{magicoder}, ODA-Fin-SFT-318k~\citep{cao2026finoda}, and Medical-Reasoning-SFT-Trinity-Mini~\citep{medical_reasoning_sft_trinity_mini}.

\paragraph{Toy derivation for repeated counting.}
Consider a linear average of \(T\) adapters. Suppose each update contains one shared unit direction \(u_0\) with coefficient \(a\), and one task-specific unit direction \(u_t\) with coefficient \(b\), where \(u_0,u_1,\ldots,u_T\) are orthonormal:
\[
\Delta W_t = a u_0 + b u_t .
\]
The averaged update is
\[
\Delta W_{\mathrm{avg}} =
\frac{1}{T}\sum_{t=1}^T \Delta W_t
= a u_0 + \frac{b}{T}\sum_{t=1}^T u_t .
\]
The shared component retains its full coefficient \(a\), while each task-specific component is reduced to \(b/T\). The shared-to-specific ratio therefore grows by a factor of \(T\). The overall norm also drops from \(\sqrt{a^2+b^2}\) per source adapter to \(\sqrt{a^2+b^2/T}\) for the merged update, which motivates the magnitude rescaling in Pico (Section~\ref{sec:pico-merge}). This derivation does not imply that every shared direction is harmful; it only shows why directions reused by many adapters can dominate a linear merge.

\paragraph{Subspace overlap.}
For a pair of adapters $i$ and $j$, let $Q_i^B$ and $Q_j^B$ be orthonormal bases for the column spaces of $B_i$ and $B_j$, and let $Q_i^A$ and $Q_j^A$ be orthonormal bases for the row spaces of $A_i$ and $A_j$. The normalized overlap scores are
\[
O_B(i,j) = \frac{1}{r}\left\| (Q_i^B)^\top Q_j^B \right\|_F^2,
\qquad
O_A(i,j) = \frac{1}{r}\left\| (Q_i^A)^\top Q_j^A \right\|_F^2.
\]
Both scores lie in $[0,1]$, and larger values mean that the two adapters share more directions. We report both the layerwise values and their averages across domain pairs.

\paragraph{Effective rank.}
Given singular values $\{\sigma_k\}$ of a matrix, the effective rank is $\mathrm{EffRank}=\exp\!\bigl(-\sum_k p_k \log p_k\bigr)$ with $p_k=\sigma_k/\sum_\ell \sigma_\ell$. A lower effective rank means that most of the singular-value mass is concentrated in fewer directions.

\paragraph{Component energy.}
For a fixed decomposition with component coefficients $\{c_j\}$, we define the energy of component $j$ as the normalized squared coefficient
\[
e_j = \frac{c_j^2}{\sum_k c_k^2}.
\]
For singular components, $c_j=\sigma_j$. Intuitively, component energy is the fraction of the total squared magnitude carried by one component; high energy means that the decomposition is dominated by that component.

Table~\ref{tab:motivation-main-stats} gives the exact values behind the two trends in Figure~\ref{fig:motivation}. The gap between $O_B$ and $O_A$ grows steadily with LoRA rank, and the effective rank of $B$ stays much lower than that of $A$ throughout. The asymmetry is already almost universal at rank 8 and becomes universal from rank 16 onward.

\begin{table*}[h]
    \centering
    \footnotesize
    \caption{\textbf{Exact statistics underlying Figure~\ref{fig:motivation}.} $\mathrm{Frac}[O_B > O_A]$ denotes the fraction of layer pairs for which overlap on the output-side matrix exceeds overlap on the input-side matrix.}
    \label{tab:motivation-main-stats}
    \begin{tabular}{llcccccc}
        \toprule
        Rank & Module & Mean $O_B$ & Mean $O_A$ & Gap & $\mathrm{Frac}[O_B > O_A]$ & EffRank$(B)$ & EffRank$(A)$ \\
        \midrule
        8  & Query projection & 0.0166 & 0.0035 & 0.0131 & 0.995 & 2.431 & 4.112 \\
        8  & Value projection & 0.0136 & 0.0071 & 0.0065 & 0.953 & 2.274 & 3.564 \\
        \addlinespace[1pt]
        16 & Query projection & 0.0341 & 0.0055 & 0.0286 & 1.000 & 2.715 & 5.155 \\
        16 & Value projection & 0.0234 & 0.0089 & 0.0145 & 1.000 & 2.413 & 4.290 \\
        \addlinespace[1pt]
        32 & Query projection & 0.0571 & 0.0095 & 0.0476 & 1.000 & 2.859 & 5.987 \\
        32 & Value projection & 0.0433 & 0.0126 & 0.0307 & 1.000 & 2.529 & 4.845 \\
        \addlinespace[1pt]
        64 & Query projection & 0.0839 & 0.0172 & 0.0667 & 1.000 & 2.944 & 6.552 \\
        64 & Value projection & 0.0801 & 0.0196 & 0.0604 & 1.000 & 2.586 & 5.302 \\
        \bottomrule
    \end{tabular}
\end{table*}

\begin{figure*}[h]
    \centering
    \includegraphics[width=\textwidth]{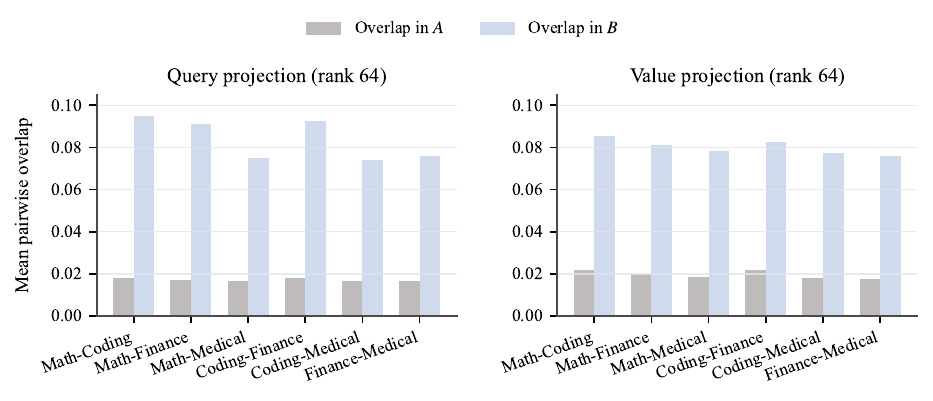}
    \caption{\textbf{Pairwise overlap at rank 64.} The asymmetry between $B$ and $A$ holds for every domain pair, showing that the effect is not driven by averaging across tasks.}
    \label{fig:motivation-pairwise-r64}
\end{figure*}

Figure~\ref{fig:motivation-pairwise-r64} complements the averaged curves in the main text with pair-level evidence. At rank 64, every domain pair satisfies $O_B > O_A$ in both the query and value projection matrices. This shows that the asymmetry is not driven by one unusually similar pair of tasks, but is shared across the full set of domain combinations.

Tables~\ref{tab:motivation-pairwise-q} and~\ref{tab:motivation-pairwise-v} provide the corresponding exact values. They make it easy to verify that the same pattern holds in both modules and across all LoRA ranks, not only at the averaged level reported in the main text.

\begin{table*}[h]
    \centering
    \footnotesize
    \caption{\textbf{Exact pair-level overlap statistics for the query projection.} Each entry reports the mean overlap for one domain pair at a fixed LoRA rank.}
    \label{tab:motivation-pairwise-q}
    \resizebox{\textwidth}{!}{%
    \begin{tabular}{lcccccccccccc}
        \toprule
        & \multicolumn{3}{c}{$r=8$} & \multicolumn{3}{c}{$r=16$} & \multicolumn{3}{c}{$r=32$} & \multicolumn{3}{c}{$r=64$} \\
        \cmidrule(lr){2-4} \cmidrule(lr){5-7} \cmidrule(lr){8-10} \cmidrule(lr){11-13}
        Domain pair & $O_B$ & $O_A$ & Gap & $O_B$ & $O_A$ & Gap & $O_B$ & $O_A$ & Gap & $O_B$ & $O_A$ & Gap \\
        \midrule
        Math--Coding    & 0.0197 & 0.0040 & 0.0157 & 0.0413 & 0.0061 & 0.0353 & 0.0663 & 0.0102 & 0.0561 & 0.0947 & 0.0179 & 0.0768 \\
        Math--Finance   & 0.0209 & 0.0034 & 0.0175 & 0.0403 & 0.0056 & 0.0347 & 0.0656 & 0.0095 & 0.0561 & 0.0913 & 0.0173 & 0.0740 \\
        Math--Medical   & 0.0131 & 0.0035 & 0.0097 & 0.0278 & 0.0053 & 0.0225 & 0.0479 & 0.0093 & 0.0385 & 0.0750 & 0.0168 & 0.0582 \\
        Coding--Finance & 0.0201 & 0.0040 & 0.0161 & 0.0413 & 0.0060 & 0.0352 & 0.0667 & 0.0102 & 0.0565 & 0.0928 & 0.0180 & 0.0747 \\
        Coding--Medical & 0.0119 & 0.0029 & 0.0090 & 0.0260 & 0.0049 & 0.0211 & 0.0472 & 0.0089 & 0.0383 & 0.0738 & 0.0167 & 0.0571 \\
        Finance--Medical& 0.0135 & 0.0029 & 0.0107 & 0.0280 & 0.0051 & 0.0230 & 0.0491 & 0.0090 & 0.0401 & 0.0760 & 0.0167 & 0.0593 \\
        \bottomrule
    \end{tabular}%
    }
\end{table*}

\begin{table*}[h]
    \centering
    \footnotesize
    \caption{\textbf{Exact pair-level overlap statistics for the value projection.} Each entry reports the mean overlap for one domain pair at a fixed LoRA rank.}
    \label{tab:motivation-pairwise-v}
    \resizebox{\textwidth}{!}{%
    \begin{tabular}{lcccccccccccc}
        \toprule
        & \multicolumn{3}{c}{$r=8$} & \multicolumn{3}{c}{$r=16$} & \multicolumn{3}{c}{$r=32$} & \multicolumn{3}{c}{$r=64$} \\
        \cmidrule(lr){2-4} \cmidrule(lr){5-7} \cmidrule(lr){8-10} \cmidrule(lr){11-13}
        Domain pair & $O_B$ & $O_A$ & Gap & $O_B$ & $O_A$ & Gap & $O_B$ & $O_A$ & Gap & $O_B$ & $O_A$ & Gap \\
        \midrule
        Math--Coding    & 0.0158 & 0.0106 & 0.0052 & 0.0261 & 0.0122 & 0.0139 & 0.0476 & 0.0155 & 0.0321 & 0.0853 & 0.0219 & 0.0634 \\
        Math--Finance   & 0.0135 & 0.0071 & 0.0064 & 0.0233 & 0.0087 & 0.0146 & 0.0444 & 0.0127 & 0.0317 & 0.0803 & 0.0198 & 0.0604 \\
        Math--Medical   & 0.0135 & 0.0060 & 0.0075 & 0.0236 & 0.0076 & 0.0159 & 0.0452 & 0.0113 & 0.0309 & 0.0784 & 0.0184 & 0.0601 \\
        Coding--Finance & 0.0144 & 0.0088 & 0.0056 & 0.0250 & 0.0112 & 0.0138 & 0.0451 & 0.0151 & 0.0300 & 0.0824 & 0.0218 & 0.0606 \\
        Coding--Medical & 0.0118 & 0.0054 & 0.0064 & 0.0210 & 0.0074 & 0.0137 & 0.0406 & 0.0108 & 0.0298 & 0.0774 & 0.0181 & 0.0593 \\
        Finance--Medical& 0.0125 & 0.0046 & 0.0079 & 0.0216 & 0.0064 & 0.0152 & 0.0399 & 0.0102 & 0.0297 & 0.0759 & 0.0176 & 0.0583 \\
        \bottomrule
    \end{tabular}%
    }
\end{table*}





\section{Additional Training Details}
\label{app:training-details}

All four domain-specific adapters are trained with the same distributed setup on 4$\times$H800 GPUs. We use a per-device batch size of 2, gradient accumulation of 8, and train for 3 epochs with AdamW optimization, bf16, gradient checkpointing, and a cosine learning-rate schedule. The learning rate is $2\times 10^{-4}$ with a warmup ratio of 0.03. We set the maximum sequence length to 8192, disable packing, apply LoRA to \texttt{q\_proj} and \texttt{v\_proj}, and use LoRA dropout 0.05. The LoRA rank varies by experiment, while LoRA alpha is fixed at 16 throughout the paper.

\section{Benchmark-Level Results}
\label{app:benchmark-level-results}
The main text reports domain averages to keep the comparison focused on the main message. This appendix gives the corresponding benchmark-level values for the two main experiment tables.

Table~\ref{tab:main-results-full} shows that the gains in the main text are not driven by a single benchmark. Pico improves the strongest downstream mergers across multiple benchmarks within each domain, with especially clear gains for Task Arithmetic on MATH and MBPP, for TIES on MATH, MBPP, and MedQA-USMLE, and for TSV-M on MATH, HumanEval, MBPP, FinanceBench, ConvFinQA, and PubMedQA. The benchmark-level view also makes the different baseline behaviors easier to see: TIES and TSV-M are particularly uneven without calibration, while Pico produces a more stable pattern across domains.

Table~\ref{tab:single-vs-joint-full} complements this comparison by showing why merging is needed in the first place. Single-domain LoRAs remain strongest on or near their source domains, but they do not provide the best overall multi-domain performance. The best merged adapter, Pico + Task Arithmetic, gives the highest overall average among the adapters compared here, while also remaining competitive on several individual benchmarks outside the source domain of any single specialist.

\begin{table*}[t]
\centering
\footnotesize
\setlength{\tabcolsep}{4.2pt}
\caption{\textbf{Benchmark-level main results across four domains.} This is the detailed version of Table~\ref{tab:main-results}.}
\label{tab:main-results-full}
\resizebox{\textwidth}{!}{%
\begin{tabular}{llccccccccc}
\toprule
\multirow{2}{*}{Method}
& \multirow{2}{*}{Merger Used}
& \multicolumn{2}{c}{Math}
& \multicolumn{2}{c}{Coding}
& \multicolumn{2}{c}{Finance}
& \multicolumn{2}{c}{Medical}
& \multirow{2}{*}{Average} \\
\cmidrule(lr){3-4}
\cmidrule(lr){5-6}
\cmidrule(lr){7-8}
\cmidrule(lr){9-10}
&
& GSM8K
& MATH
& HumanEval
& MBPP
& FinanceBench
& ConvFinQA
& PubMedQA
& MedQA-USMLE
& \\
\midrule
No Calibration & Task Arithmetic & 0.1626 & 0.3707 & 0.2805 & 0.2840 & 0.5067 & 0.5020 & 0.5850 & 0.5829 & 0.4093 \\
DARE           & Task Arithmetic & 0.1746 & 0.3768 & 0.2866 & 0.2996 & 0.5467 & 0.4946 & 0.5850 & 0.5852 & 0.4186 \\
DELLA          & Task Arithmetic & 0.1732 & 0.3571 & 0.2561 & 0.2763 & 0.5333 & 0.5040 & 0.5800 & 0.5915 & 0.4089 \\
KnOTS          & Task Arithmetic & 0.1656 & 0.3798 & 0.2744 & 0.2685 & 0.5200 & 0.5054 & 0.5780 & 0.5789 & 0.4088 \\
Core Space     & Task Arithmetic & 0.1600 & 0.3639 & 0.2744 & 0.2802 & 0.5000 & 0.4973 & 0.5900 & 0.5860 & 0.4065 \\
\textbf{Pico}  & Task Arithmetic & 0.1672 & 0.4458 & 0.3049 & 0.3696 & 0.5733 & 0.5128 & 0.5650 & 0.6057 & \textbf{0.4430} \\
\midrule
No Calibration & TIES            & 0.1658 & 0.3958 & 0.2988 & 0.1245 & 0.5467 & 0.5074 & 0.4810 & 0.5672 & 0.3859 \\
DARE           & TIES            & 0.1638 & 0.3563 & 0.2744 & 0.1128 & 0.5400 & 0.4859 & 0.4300 & 0.5648 & 0.3660 \\
DELLA          & TIES            & 0.1240 & 0.2616 & 0.2134 & 0.1284 & 0.5267 & 0.4234 & 0.2520 & 0.5530 & 0.3103 \\
KnOTS          & TIES            & 0.1704 & 0.4049 & 0.2988 & 0.3035 & 0.5400 & 0.5155 & 0.5800 & 0.5522 & 0.4207 \\
Core Space     & TIES            & 0.1580 & 0.3245 & 0.1829 & 0.3152 & 0.5200 & 0.3259 & 0.3540 & 0.6332 & 0.3517 \\
\textbf{Pico}  & TIES            & 0.1786 & 0.4337 & 0.2805 & 0.3268 & 0.5133 & 0.5060 & 0.5610 & 0.6622 & \textbf{0.4328} \\
\midrule
No Calibration & TSV-M           & 0.1754 & 0.3518 & 0.1585 & 0.1556 & 0.4667 & 0.4308 & 0.2500 & 0.7895 & 0.3473 \\
DARE           & TSV-M           & 0.2076 & 0.3776 & 0.1646 & 0.1595 & 0.4467 & 0.4496 & 0.3800 & 0.7918 & 0.3722 \\
DELLA          & TSV-M           & 0.1872 & 0.3685 & 0.1524 & 0.1673 & 0.4733 & 0.4590 & 0.2610 & 0.7926 & 0.3577 \\
KnOTS          & TSV-M           & 0.1846 & 0.3487 & 0.1646 & 0.1518 & 0.4733 & 0.4516 & 0.2550 & 0.7887 & 0.3523 \\
Core Space     & TSV-M           & 0.1836 & 0.2934 & 0.1585 & 0.1245 & 0.4400 & 0.2735 & 0.1190 & 0.7636 & 0.2945 \\
\textbf{Pico}  & TSV-M           & 0.1740 & 0.4026 & 0.2927 & 0.2957 & 0.5400 & 0.5114 & 0.5720 & 0.6559 & \textbf{0.4305} \\
\bottomrule
\end{tabular}%
}
\end{table*}

\begin{table*}[t]
\centering
\footnotesize
\setlength{\tabcolsep}{4.2pt}
\caption{\textbf{Benchmark-level comparison with single-domain and jointly trained LoRA adapters.} This is the detailed version of Table~\ref{tab:single-vs-joint}.}
\label{tab:single-vs-joint-full}
\resizebox{\textwidth}{!}{%
\begin{tabular}{lccccccccc}
\toprule
\multirow{2}{*}{Adapter}
& \multicolumn{2}{c}{Math}
& \multicolumn{2}{c}{Coding}
& \multicolumn{2}{c}{Finance}
& \multicolumn{2}{c}{Medical}
& \multirow{2}{*}{Average} \\
\cmidrule(lr){2-3}
\cmidrule(lr){4-5}
\cmidrule(lr){6-7}
\cmidrule(lr){8-9}
&
GSM8K
& MATH
& HumanEval
& MBPP
& FinanceBench
& ConvFinQA
& PubMedQA
& MedQA-USMLE
& \\
\midrule
Math LoRA             & 0.1892 & 0.3768 & 0.1402 & 0.0778 & 0.4600 & 0.4207 & 0.2930 & 0.4399 & 0.2997 \\
Coding LoRA           & 0.1728 & 0.3283 & 0.3110 & 0.4086 & 0.4867 & 0.4758 & 0.5510 & 0.5522 & 0.4108 \\
Finance LoRA          & 0.1640 & 0.3313 & 0.1829 & 0.1051 & 0.5667 & 0.5302 & 0.2420 & 0.7533 & 0.3594 \\
Medical LoRA          & 0.1660 & 0.3215 & 0.2683 & 0.3696 & 0.5333 & 0.4362 & 0.4170 & 0.7730 & 0.4106 \\
\midrule
Joint Multi-Task LoRA & 0.1774 & 0.3313 & 0.3110 & 0.3191 & 0.6333 & 0.2977 & 0.2650 & 0.6159 & 0.3688 \\
\textbf{Pico w/ Task Arithmetic}
                      & 0.1672 & 0.4458 & 0.3049 & 0.3696 & 0.5733 & 0.5128 & 0.5650 & 0.6057 & \textbf{0.4430} \\
\bottomrule
\end{tabular}%
}
\end{table*}

\section{Definitions and Full Spectral Statistics of the Merged \texorpdfstring{$B$}{B} Space}
\label{app:bspace-spectrum}

This appendix defines the spectral diagnostics used in Figure~\ref{fig:bspace-stats} and reports their exact values. These diagnostics let us check whether the merged output-side matrix $B$ is dominated by a few over-shared components or spreads its energy more evenly across the shared basis.

We use four diagnostics to summarize the spectrum of a merged $B$ matrix. The first is $o_{\max}$, the largest normalized energy share among the singular components. A lower $o_{\max}$ means that no single shared component dominates the spectrum. The second is effective rank, which measures how evenly the singular-value mass is distributed. The third is stable rank, which plays a similar role: higher effective rank and higher stable rank both indicate that energy is spread across more components instead of concentrating in only a few. We also report the condition number as an additional measure of spectral skewness, where lower values are better. Finally, Frobenius norm is included only as a reference magnitude and does not have a preferred direction.

For completeness, if $\{\sigma_j\}$ are the singular values of the merged $B$ matrix, then the normalized energy share of component $j$ is $o_j=\sigma_j^2 / \sum_k \sigma_k^2$, and $o_{\max}$ is the largest such value. Effective rank is computed from the normalized singular values, and stable rank is $\|B\|_F^2 / \|B\|_2^2$.


\begin{table*}[t]
\centering
\footnotesize
\setlength{\tabcolsep}{6.2pt}
\caption{\textbf{Full spectral statistics of the merged $B$ space.} Frobenius norm is reported as a reference magnitude. Lower $o_{\max}$ and condition number are better; higher effective rank and stable rank are better.}
\label{tab:bspace-stats-full}
\resizebox{\textwidth}{!}{%
\begin{tabular}{lccccc}
\toprule
Adapter & Frobenius Norm (ref.) & $o_{\max}\downarrow$ & Effective Rank $\uparrow$ & Stable Rank $\uparrow$ & Condition Number $\downarrow$ \\
\midrule
\multicolumn{6}{l}{\textit{Merged adapters}} \\
\midrule
\textbf{Pico}  & 2.4547 & 0.2192 & 11.4733 & 4.7581 & 4.0219 \\
No Calibration & 1.7220 & 0.3268 & 9.2866 & 3.2293 & 5.8863 \\
DARE           & 1.7247 & 0.3249 & 9.3698 & 3.2438 & 5.5169 \\
KnOTS          & 1.7220 & 0.3268 & 9.2868 & 3.2292 & 5.8683 \\
\midrule
\multicolumn{6}{l}{\textit{Source-domain adapters}} \\
\midrule
Math LoRA      & 1.1537 & 0.7699 & 2.3481 & 1.3302 & 33.7237 \\
Finance LoRA   & 1.5313 & 0.5859 & 4.2613 & 1.8803 & 20.4689 \\
Medical LoRA   & 1.2375 & 0.7263 & 2.6762 & 1.4430 & 29.4735 \\
Coding LoRA    & 1.1778 & 0.7477 & 2.5098 & 1.4018 & 31.4788 \\
\bottomrule
\end{tabular}%
}
\end{table*}

\section{Additional Rank-Robustness Results}
\label{app:rank-robustness}

This appendix gives the full rank-sweep results that supplement the short discussion in Section~\ref{sec:exp-ablation}. The goal is to check whether the advantage of calibrating $B$ before merge remains visible when the LoRA rank changes.

Table~\ref{tab:rank-robustness} shows that the pattern in the main text continues across all four ranks. Pico remains the strongest method for Task Arithmetic, TIES, and TSV-M at ranks 8, 16, 32, and 64. The gap is especially clear for TSV-M, whose uncalibrated performance drops sharply at higher ranks, while Pico stays clearly stronger throughout the full rank sweep.

\begin{table*}[t]
\centering
\footnotesize
\setlength{\tabcolsep}{6.0pt}
\caption{\textbf{Robustness across LoRA ranks.} Each entry reports the overall average across all eight benchmarks. Pico remains the strongest method for all three downstream mergers at ranks 8, 16, 32, and 64.}
\label{tab:rank-robustness}
\resizebox{\textwidth}{!}{%
\begin{tabular}{c c c c c c c c}
\toprule
Rank & Merge Method & No Calibration & DARE & DELLA & KnOTS & Core Space & Pico \\
\midrule
\multirow{3}{*}{8}
  & Task Arithmetic & 0.4285 & 0.4404 & 0.4473 & 0.4412 & 0.4662 & \textbf{0.4890} \\
  & TIES            & 0.4816 & 0.4857 & 0.4353 & 0.4997 & 0.4395 & \textbf{0.5008} \\
  & TSV-M           & 0.3105 & 0.2811 & 0.2790 & 0.2957 & 0.2807 & \textbf{0.4847} \\
\midrule
\multirow{3}{*}{16}
  & Task Arithmetic & 0.4093 & 0.4186 & 0.4089 & 0.4088 & 0.4065 & \textbf{0.4430} \\
  & TIES            & 0.3859 & 0.3660 & 0.3103 & 0.4207 & 0.3517 & \textbf{0.4328} \\
  & TSV-M           & 0.3473 & 0.3722 & 0.3577 & 0.3523 & 0.2945 & \textbf{0.4305} \\
\midrule
\multirow{3}{*}{32}
  & Task Arithmetic & 0.4754 & 0.4679 & 0.4505 & 0.4823 & 0.4802 & \textbf{0.5279} \\
  & TIES            & 0.5005 & 0.4827 & 0.4748 & 0.5029 & 0.4801 & \textbf{0.5043} \\
  & TSV-M           & 0.3206 & 0.3031 & 0.3096 & 0.3173 & 0.2977 & \textbf{0.4621} \\
\midrule
\multirow{3}{*}{64}
  & Task Arithmetic & 0.4257 & 0.4504 & 0.4400 & 0.4270 & 0.4633 & \textbf{0.5221} \\
  & TIES            & 0.4858 & 0.4878 & 0.4640 & 0.4870 & 0.4749 & \textbf{0.4941} \\
  & TSV-M           & 0.2926 & 0.3052 & 0.2919 & 0.2909 & 0.3011 & \textbf{0.3976} \\
\bottomrule
\end{tabular}%
}
\end{table*}

\section{Additional Results on Qwen3-4B-Base}
\label{app:qwen-base-transfer}

This appendix reports the base-model transfer result, summarized briefly in Section~\ref{sec:exp-ablation}. Table~\ref{tab:qwen-base-transfer} repeats the Task Arithmetic comparison on Qwen3-4B-Base under the same four-domain setup and shows that Pico again gives the best overall average.

Table~\ref{tab:qwen-base-transfer} shows that the overall pattern from the main text survives the backbone change. Pico again gives the best overall average, with especially large gains in math and medical, while coding remains strong instead of dropping sharply.

\begin{table*}[t]
\centering
\small
\setlength{\tabcolsep}{4.0pt}
\caption{\textbf{Transfer to Qwen3-4B-Base.} Domain averages are computed from the same eight benchmarks as in the main experiment. Pico again gives the best overall average.}
\label{tab:qwen-base-transfer}
\resizebox{\textwidth}{!}{%
\begin{tabular}{lccccc}
\toprule
Method & Math Avg. $\uparrow$ & Coding Avg. $\uparrow$ & Finance Avg. $\uparrow$ & Medical Avg. $\uparrow$ & Overall Avg. $\uparrow$ \\
\midrule
No Calibration & 0.1952 & 0.5798 & 0.6144 & 0.5632 & 0.4881 \\
DARE           & 0.2607 & 0.5718 & 0.6167 & 0.5539 & 0.5008 \\
DELLA          & 0.2326 & 0.5737 & 0.5996 & 0.5574 & 0.4908 \\
KnOTS          & 0.2622 & 0.5593 & 0.6113 & 0.5661 & 0.4997 \\
Core-TA        & 0.1264 & 0.5563 & 0.5515 & 0.5187 & 0.4382 \\
\textbf{Pico}  & \textbf{0.3265} & \textbf{0.5825} & \textbf{0.6545} & \textbf{0.6594} & \textbf{0.5557} \\
\bottomrule
\end{tabular}%
}
\end{table*}

\section{Progressive LoRA Merging on Coding}
\label{app:progressive-coding}

\begin{figure*}[htbp]
\centering
\includegraphics[width=.86\textwidth]{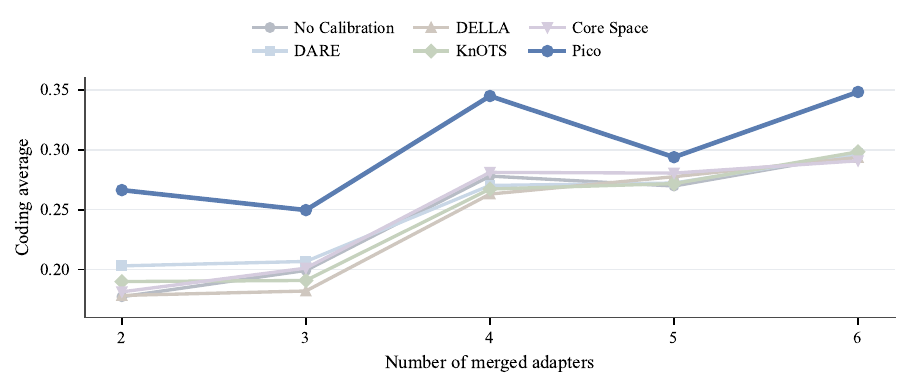}
\caption{\textbf{Progressive LoRA merging on coding benchmarks.} The x-axis shows the number of merged adapters. Starting from the coding adapter, we progressively add finance, medical, math, and law adapters and report the coding average after each merge. Pico remains the strongest method at every stage.}
\label{fig:progressive-coding}
\end{figure*}

This appendix studies a progressive merge protocol in which the adapter pool grows over time. We start from the coding adapter and then add the finance, medical, math, and law adapters one at a time, evaluating the coding average after each stage. The law adapter is trained on Lawyer-Instruct~\citep{law}. Figure~\ref{fig:progressive-coding} reports the resulting coding average as the number of merged adapters increases from two to six.

Figure~\ref{fig:progressive-coding} shows that the coding curve is not monotonic. Adding another adapter can hurt coding, but it can also help. A new adapter may introduce extra interference, and it may also bring in structure that transfers well to coding. That is why some baselines improve when the merge grows from three adapters to four instead of dropping at every step.

The main pattern is that the baseline methods become less stable as more independently trained updates are merged, while Pico remains strongest at every stage. This matches the main story of the paper. With more adapters in the merge, the same shared $B$ directions are more likely to be counted repeatedly across tasks. Pico calibrates those directions before merge, which makes coding performance more stable in this harder setting.



\end{document}